%% file: dbcnn_cvpr18.tex
\documentclass[10pt,twocolumn,letterpaper]{article}

\usepackage{cvpr}
\usepackage{times}
\usepackage{epsfig}
\usepackage{graphicx}
\usepackage{amsmath}
\usepackage{amssymb}
\usepackage{adjustbox,lipsum}
\usepackage[vlined,boxed,commentsnumbered]{algorithm2e}
\usepackage{subfigure}
\usepackage{tablefootnote}

\usepackage[pagebackref=true,breaklinks=true,letterpaper=true,colorlinks,bookmarks=false]{hyperref}

\cvprfinalcopy 

\def\cvprPaperID{329} 

\input{definitions}

\ifcvprfinal\fi
\begin{document}

\title{Video Representation Learning Using Discriminative Pooling}

\author{
Jue Wang${^{1,3}}$\quad Anoop Cherian$^{2,3,4}$\quad Fatih Porikli${^{2,3}}$\quad Stephen Gould${^{2,3}}$ \\
 ${^1}$Data61/CSIRO,\quad ${^2}$Australian Centre for Robotic Vision \\
 ${^3}$The Australian National University, Canberra, Australia, ${^4}$MERL Cambridge, MA\\
{\tt\small firstname.lastname@anu.edu.au}
}
\maketitle
\input{abstract}
\input{intro}
\input{related_work}
\input{proposed_method}

\input{end2end}

\input{expts}

\input{conclude}

{\small
\bibliographystyle{ieee}
\bibliography{dbcnn_cvpr}
}

\end{document}

%% file: definitions.tex
\newtheorem{definition}{Definition}

\newcommand{\reals}[1]{\mathbb{R}^{#1}}
\newcommand{\enorm}[1]{\left\|{#1}\right\|}

\newcommand{\set}[1]{\left\{#1\right\}}

\DeclareMathOperator*{\subjectto}{\text{subject to}}
\newcommand{\eye}[1]{\mathbf{I}_{#1}}

\renewcommand\cdots{...}

\newcommand{\suptensor}[1]{\mathfrak{S}^{d}}

\DeclareMathOperator*{\argmin}{arg\,min}

\newcommand{\comment}[1]{}

\newcommand{\done}[1]{}
\newcommand{\actodo}[1]{}

\newcommand{\seq}{X}
\newcommand{\idx}{\text{I}}
\newcommand*{\pseq}[1]{X^{\small{+}}_{{#1}}}
\newcommand*{\hpseq}[1]{\hat{X}^{\small{+}}_{{#1}}}
\newcommand*{\nseq}[1]{X^{\small{-}}_{{#1}}}
\newcommand*{\pdataset}{\mathcal{X}^{\small{+}}_{}}
\newcommand*{\ndataset}{\mathcal{X}^{\small{-}}_{}}
\newcommand*{\nfeat}[2]{\mathbf{x}^{{#1}\small{-}}_{{#2}}}
\newcommand*{\pfeat}[2]{\mathbf{x}^{{#1}\small{+}}_{{#2}}}

\newcommand*{\feat}{\mathbf{x}}
\newcommand*{\hfeat}{\mathbf{\hat{x}}}

\newcommand*{\ypseq}[1]{y^{\small{+}}_{{#1}}}

\newcommand{\card}[1]{\left|{#1}\right|}
\DeclareMathOperator*{\svmp}{SVMP}

\DeclareMathOperator*{\svm}{SVM}
\newcommand{\dataset}{\mathcal{X}}

\newcommand{\wbstack}{\left[\begin{array}{c}\!\!\!w\!\!\!\\\!\!\!b\!\!\!\end{array}\right]}

%% file: abstract.tex
\begin{abstract}
Popular deep models for action recognition in videos generate independent predictions for short clips, which are then pooled heuristically to assign an action label to the full video segment. As not all frames may characterize the underlying action---indeed, many are common across multiple actions---pooling schemes that impose equal importance on all frames might be unfavorable. In an attempt to tackle this problem, we propose~\emph{discriminative pooling}, based on the notion that among the deep features generated on all short clips, there is at least one that characterizes the action. To this end, we learn a (nonlinear) hyperplane that separates this unknown, yet discriminative, feature from the rest. Applying multiple instance learning in a large-margin setup, we use the parameters of this separating hyperplane as a descriptor for the full video segment. Since these parameters are directly related to the support vectors in a max-margin framework, they serve as robust representations for pooling of the features. We formulate a joint objective and an efficient solver that learns these hyperplanes per video and the corresponding action classifiers over the hyperplanes. Our pooling scheme is end-to-end trainable within a deep framework. We report results from experiments on three benchmark datasets spanning a variety of challenges and demonstrate state-of-the-art performance across these tasks.



\end{abstract}

%% file: intro.tex
\section{Introduction}
\label{sec:intro}
We are witnessing an astronomical increase of video data on the web. This data deluge has brought out the problem of effective video representation -- specifically, their semantic content -- to the forefront of computer vision research. The resurgence of convolutional neural networks (CNN) has enabled significant progress to be made on several problems in computer vision (most notably on object detection and image tagging) and is now pushing forward the state-of-the-art in action recognition and video understanding. However, current solutions are still far from being practically useful, arguably due to the volumetric nature of this data modality and the complex nature of real-world human actions~\cite{feichtenhofer2016spatiotemporal,feichtenhofer2017spatiotemporal,feichtenhofer2017temporal,feichtenhofer2016convolutional,hayat2015deep, simonyan2014two, simonyan2014very,Wang2016}. 
 
\begin{figure}
	\begin{center}
        \includegraphics[width=0.8\linewidth,trim={0cm 0cm 0cm 0cm},clip]{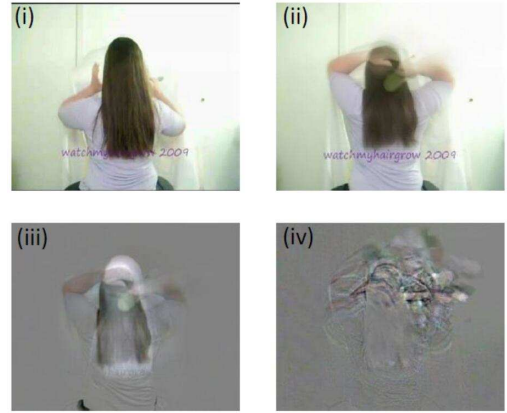}
	\end{center}
	\caption{A visualization of discriminative pooling applied to RGB frames in a sequence. (i) a sample frame, (ii) average pooling all frames, (iii) dynamic image by rank pooling~\cite{bilen2016dynamic}, and (iv) our SVM pooling. Our representation captures more details of the actions as it learns to discriminate parts of the foreground against a background set. In this case, we used the average-pooled frames as the background. }
	\label{fig:1}
\end{figure}

Using effective architectures, CNNs are often found to extract features from images that perform well on recognition tasks. Leveraging this know-how, deep learning solutions for video action recognition have so far been straightforward extensions of image-based models. However, video data can be of arbitrary length and scaling up image-based CNN architectures to yet another dimension of complexity is not an easy task as the number of parameters in such models will be significantly higher. This demands more advanced computational infrastructures and greater quantities of clean training data~\cite{carreira2017quo}. Instead, the trend has been on converting the video data to short temporal segments consisting of one to a few frames, on which the existing image-based CNN models are trained. For example, in the popular two-stream model~\cite{feichtenhofer2016convolutional, simonyan2014two, simonyan2014very, wang2015action, wangtwo}, the CNNs are trained to independently predict actions from short video clips (consisting of single frames or stacks of about ten optical flow frames); these predictions are then pooled to generate a prediction for the full sequence -- typically using average/max pooling or classifying using a linear SVM. While average pooling gives equal weights to all the predictions, max pooling is sensitive to outliers, and a classifier may be confused by predictions from background actions. Several works try to tackle this problem by using different pooling strategies \cite{bilen2016dynamic, fernando2015modeling, wang2015action,Wang2016,yue2015beyond}, which achieve some improvement compared with the baseline algorithm. 

To this end, we observe that not all predictions on the short video snippets are equally informative, yet some of them must be~\cite{schindler2008action}. This allows us to cast the problem in a multiple instance learning (MIL) framework, where we assume that some of the features in one sequence are indeed useful, while the rest are not. We assume all the CNN features from a sequence (containing both the good and bad features) to represent a positive bag, while features from the known background or noisy frames as a negative bag. We then formulate a binary classification problem of separating as many good features as possible in the positive bag using a discriminative classifier. The decision boundary of the classifier thus learned is then used as a descriptor for the entire video sequence, which we call the SVM Pooled (SVMP) descriptor. Subsequently, the SVMP descriptor is used in an action classification setup. We also provide a joint objective that learns both the SVMP descriptors and the action classifiers, which generalizes the applicability of our method to both CNN features and hand-crafted ones. In a pure CNN setup, our pooling method can be implemented alongside the rest of the CNN layers and trained in an end-to-end manner.

Compared to other popular pooling schemes, our proposed method offers several benefits: First, it produces a compact representation of a video sequence of arbitrary length by characterizing the classifiability of its features against a background set. Second, it is robust to classifier outliers thanks to the SVM formulation. Last, it is computationally efficient. To provide intuitions, in Figure~\ref{fig:1}, we provide a visualization of our descriptor applied directly on video frames. As is clear, SVMP captures the essence of action dynamics in more detail in comparison to prior works.

We provide extensive experimental evidence on various datasets for different tasks, such as action recognition/anticipation/detection on the HMDB-51 and Charades datasets and skeleton-based action recognition on NTU-RGBD. We outperform baseline results on these datasets by a significant margin (between 3--11\%) and beat all the previously reported results as well (between 1--4\%).

To set the stage for introducing our method, we briefly review relevant literature on prior works in the next section.

%% file: related_work.tex
\section{Related Work}
\label{sec:related_work}
Traditional methods for video action recognition typically use hand-crafted local features, such as dense trajectories, HOG, HOF, etc.~\cite{wang2013action}, or mid-level representations on them, such as Fisher Vectors~\cite{sadanand2012action}. With the resurgence of deep learning methods for object recognition~\cite{krizhevsky2012imagenet}, there have been several attempts to adapt these models to action recognition. Recent practice is to feed the video data, including RGB frames, optical flow subsequences, and 3D skeleton data into a deep (recurrent) network to train it in a supervised manner. Successful methods following this approach are the two-stream models and their extensions~\cite{feichtenhofer2017spatiotemporal,feichtenhofer2017temporal,hayat2015deep,kim2017interpretable,simonyan2014two}. Although the architecture of these networks are different, the core idea is to embed the video data into a semantic feature space, and then recognize the actions either by aggregating the individual features per frame using some statistic (such as max or average) or directly training a CNN based end-to-end classifier~\cite{feichtenhofer2017spatiotemporal}. While the latter schemes are appealing, they usually need to store the feature maps from all the frames in memory which may be prohibitive for longer sequences. This problem may be tackled using recurrent models~\cite{baccouche2011sequential,donahue2015long,du2015hierarchical,li2016action,srivastava2015unsupervised,yue2015beyond}, however such models are usually harder to train~\cite{pascanu2013difficulty}. Another promising direction is to use 3D convolutional filters~\cite{carreira2017quo,tran2015learning}, but would need more parameters and large amounts of clean data for pretraining. In contrast to all these approaches, we look at the problem from that of choosing the correct set of frames automatically that are discriminative in recognizing the actions. A scheme similar to ours is the recent work of Wang et al.,~\cite{Wang2016}, however they use manually-defined video segmentation for equally-spaced snippet sampling.

Typically, pooling schemes consolidate input data into compact representations. Instead, we use the parameters of the data modeling function, i.e., the SVM decision boundary, as our representation. Note that such a hyperplane is of the same dimensionality as the data and well-known as a weighted combination of each data point, where the weight captures how discriminative each point is. There have been other recent works that use parameters of machine learning algorithms for the purpose of pooling, such as rank pooling~\cite{fernando2015modeling}, generalized rank pooling~\cite{grp}, dynamic images~\cite{bilen2016dynamic} and dynamic flow~\cite{dynamic_flow}. However, while these methods optimize a rank-SVM based regression formulation, our motivation and formulation are different. We use the parameters of a binary SVM to be the video level descriptor, which is trained to classify the frame level features from a preselected (but arbitrary) bag of negative features. In this respect, our pooling scheme is also different from Exemplar-SVMs~\cite{malisiewicz2011ensemble,willems2009exemplar,zepeda2015exemplar} that learns feature filters per data sample and then use these filters for feature extraction.


An important component of our scheme is the MIL scheme, which is a popular data selection technique~\cite{cinbis2017weakly,li2015multiple,wu2015deep,yi2016human,zhang2015self}. In the context of action recognition, schemes similar in motivation have been suggested before. For example, Satkin and Hebert~\cite{satkin2010modeling} explore the effect of temporal cropping of videos to regions of actions; however, assumes these regions are continuous. Nowozin et al.~\cite{nowozin2007discriminative} represent videos as sequences of discretized spatio-temporal sets and reduces the recognition task into a max-gain sequence finding problem on these sets using an LPBoost classifier. Similar to ours, Li et al.~\cite{li2013dynamic} propose an MIL setup for complex activity recognition using a dynamic pooling operator--a binary vector that selects input frames to be part of an action, which is learned by reducing the MIL problem to a set of linear programs. Chen and Nevatia~\cite{sun2014discover} propose a latent variable based model to explicitly localize discriminative video segments where events take place. Vahdat et al. present a compositional model in ~\cite{vahdat2013compositional} for video event detection using a multiple kernel learning based latent SVM. While all these schemes share similar motivations as ours, we cast our MIL problem in the setting of normalized set kernels~\cite{gartner2002multi} and reduce the formulation to standard SVM setup which can be solved rapidly. In the $\propto$-SVMs of Yu et al.,~\cite{lai2014video,yu2013propto}, the positive bags are assumed to have a fixed fraction of positives, which is a criterion we also assume in our framework. However, our optimization setup and our goals are different. Specifically, our goal is to learn a video representation for recognition, while~\cite{yu2013propto} tackles the problem of action detection. 

\begin{figure*}[ht]
	\begin{center}
        \includegraphics[width=0.85\linewidth,trim={0cm 0cm 0cm 0cm},clip]{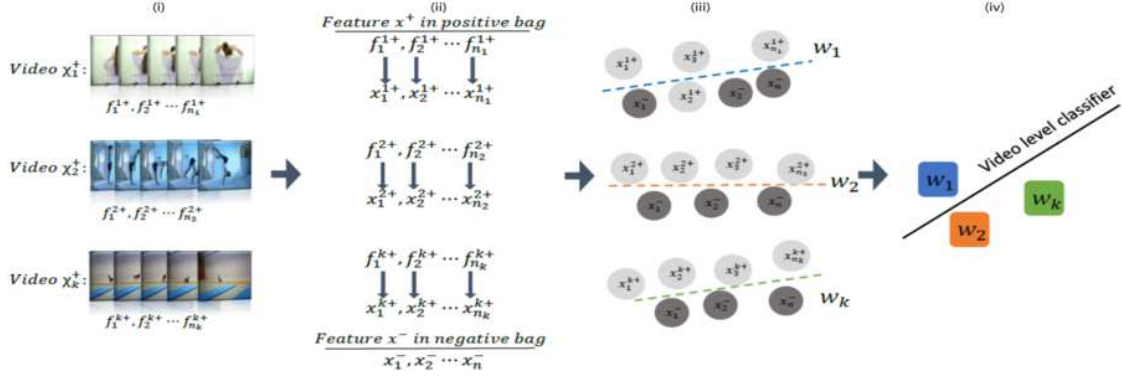}
	\end{center}
	\caption{Illustration of our SVM pooling pipeline: (i) Extraction of frames $f_i$ from videos, (ii) converting frames $f_i$ into features $x_i$, (iii) learning decision boundaries $w_j$, one for every sequence, on its respective features $x_i$, and (iv) using $w_j$ as descriptor in a video classifier.} 
   	\label{fig:2}
\end{figure*}

%% file: proposed_method.tex
\section{Proposed Method}
\label{sec:setup}
In this section, we describe our method for learning SVMP descriptors and the action classifiers. The overall pipeline is illustrated in Figure \ref{fig:2}. 

Let us assume we are given a dataset of $N$ video sequences $\pdataset = \set{\pseq{1}, \pseq{2},\cdots, \pseq{N}}$, where each $\pseq{i}$ is sequence with a set of frame level features, $\ie$, $\pseq{i}=\set{\pfeat{i}{1}, \pfeat{i}{2}, \cdots, \pfeat{i}{n}}$, each $\pfeat{i}{k}\in\reals{p}$. We assume that each $\pseq{i}$ is associated with an action class label $\ypseq{i}\in\set{1,2,\cdots, d}$. Further, the $+$ sign denotes that the features and the sequences represent a positive bag. We also assume that we have access to a set of sequences $\ndataset=\set{\nseq{1}, \nseq{2},\cdots \nseq{M}}$ belonging to actions different from those in $\pdataset$, where each $\nseq{j}=\set{\nfeat{j}{1}, \nfeat{j}{2}, \cdots, \nfeat{j}{n}}$ are the features associated with a negative bag, each $\nfeat{j}{k}\in\reals{p}$. For simplicity, we assume all sequences have the same number $n$ of features. Further our scheme is feature-agnostic, i.e., they may be from a CNN or are hand-crafted. 

Our goals are two-fold, namely (i) to learn a classifier decision boundary for every sequence in $\pdataset$ that separates a fraction $\eta$ of them from the features in $\ndataset$ and (ii) to learn video level classifiers on the classes in the positive bags that are represented by the learned decision boundaries in (i). We provide below an MIL formulation for (i) and a joint objective combining (i) and learning (ii).

\subsection{Learning Decision Boundaries}

As described above, our goal in this section is to generate a descriptor for each sequence $\pseq{}\in\pdataset$; this descriptor we define to be the learned parameters of a hyperplane that separates the features $\pfeat{}{}\in\pseq{}$ from all features in $\ndataset$. We do not want to warrant that all $\pfeat{}{}$ can be separated from $\ndataset$ (as several of them may belong to a background class), however we assume that at least a fixed fraction $\eta$ of them are classifiable. Mathematically, suppose the tuple $(w_i,b_i)$ represents the parameters of a max-margin hyperplane separating some of the features in a positive bag $\pseq{i}$ from all features in $\ndataset$, then we cast the following objective, which is a variant of the sparse MIL (SMIL)~\cite{bunescu2007multiple}, normalized set kernel (NSK)~\cite{gartner2002multi}, and $\propto$-SVM~\cite{yu2013propto} formulations:
\begin{align}
\label{eq:mil}
&\!\!\argmin_{w_i\in\reals{p},b_i\in\reals{},\zeta\geq 0} P1(w_i,b_i) := \enorm{w_i}^2 +  C_1\sum_{k=1}^{(M+1)n}\zeta_k\\
&\subjectto\ \theta(\feat;\eta)\left(w_i^T\feat+b_i\right)  \geq 1 - \zeta_k\\
\label{eq:5}&\theta(\feat;\eta)= -1, \forall \feat \in \left\{\pseq{i}\bigcup \ndataset\right\}\backslash \hpseq{i}\\
\label{eq:6}&\theta(\hfeat; \eta) = 1, \forall \hfeat \in\hpseq{i}  \\
&\card{\hpseq{i}}  \geq \eta \card{\pseq{i}}. \quad\quad
\label{eq:ratio-constraint} 
\end{align} 
In the above formulation, we assume that there is a subset $\hpseq{i}\subset\pseq{i}$ that is classifiable, while the rest of the positive bag need not be, as captured by the ratio in~\eqref{eq:ratio-constraint}. The variables $\zeta$ capture the non-negative slacks weighted by a regularization parameter $C_1$, and the function $\theta$ provides the label of the respective features. Unlike SMIL or NSK objectives, that assumes the individual features $\feat$ are summable, our problem is non-convex due to the unknown set $\hpseq{}$. However, this is not a serious deterrent to the usefulness of our formulation and can be tackled easily as described in the sequel and supported by our experimental results.

Given that the above formulation is built on an SVM objective, we call this specific discriminative pooling scheme as~\emph{SVM pooling} and formally define the descriptor for a sequence as:
\begin{definition}[SVM Pooling Desc.]
\label{def:svmp}
Given a sequence $\seq$ of features $\feat\in\reals{p}$ and a negative dataset $\ndataset$, we define the~\emph{SVM Pooling} (SVMP) descriptor as $\svmp(\seq) = [w,b]^T\in\reals{p+1}$, where the tuple $(w,b)$ is obtained as the solution of problem $P1$ defined in~\eqref{eq:mil}.
\end{definition}

\subsection{Learning Video Level Classifiers}
\label{learn_classifier}
Given a dataset of sequences $\pdataset$ and a negative bag $\ndataset$, we propose to learn the SVMP descriptors per sequence and the classifiers on $\pdataset$ jointly as a multi-class structured SVM problem which includes the MIL problem $P1$ as a sub-objective. The joint formulation is as follows:
\begin{align}
\min_{w,b,Z} P2 :=\sum_{i=1}^N & \enorm{w_i}^2 +  \sum_{j=1}^d \enorm{Z_j}^2\notag\\  &+ C_2\sum_{i,l=1}^N \gamma_{il} + C_1\sum_{i=1}^{N} \!\!\sum_{k=1}^{(M+1)n}\hspace*{-0.3cm}\zeta_{ik}
\end{align}
\vspace*{-0.5cm}
\begin{align}
\label{eq:9}
&Z_j^T\left(\wbstack_i-\wbstack_l\right) \geq \Delta(\ypseq{i},\ypseq{l})-\gamma_{il},\\
&\text{where } \ypseq{i} = j, \forall \ypseq{l}\in \idx_d, \text{ and } \forall j \in \idx_d, \notag
\end{align}
\vspace*{-0.8cm}
\begin{align}
\theta(\feat; \eta)\left(w_i^T\feat+b_i\right) &\geq 1 - \zeta_{ik},\  \feat\in\pseq{i}\bigcup\ndataset, \forall i=\idx_N,\nonumber\\ 
\text{and}\quad \theta(\feat; \eta) &\in\set{+1,-1}, \gamma_{il}\geq 0,\ \zeta_{ik}\geq 0,\nonumber
\end{align}
where $\theta(x.;\eta)$ is as defined in~\eqref{eq:5} and~\eqref{eq:6}. The function $\Delta(y,z)$ computes the similarity between the ground truth labels $y$ and $z$. The formulation $P2$ jointly optimizes the computations of SVMP descriptors per sequence $(w_i,b_i)$ and the parameters $Z$ of $d$ video level classifiers, in a one-versus-rest fashion as described in~\eqref{eq:9}. The constant $C_2$ is a regularization parameter on the action classifiers and $\gamma$ represents the respective slack variables per sequence. For brevity, we use $I_m$ to represent the set $\set{1,2,\cdots, m}$.

\subsection{Efficient Optimization}
The problem $P2$ is not convex due to the function $\theta(\feat;\eta)$ that needs to select a set from the positive bags that satisfy the criteria in~\eqref{eq:ratio-constraint}. Also, note that the sub-problem $P1$ could be posed as a mixed-integer quadratic program (MIQP), which is known to be in NP~\cite{lazimy1982mixed}. While, there are efficient approximate solutions for this problem (such as~\cite{misener2013glomiqo}), the solution must be scalable to large number of both high-dimensional features generated by a CNN and low-dimensional local features. To this end, we propose the following relaxation.

Note that the regularization parameter $C_1$ in~\eqref{eq:mil} controls the positiveness of the slack variables $\zeta$, thereby influencing the training error rate. A smaller value of $C_1$ allows more data points to be misclassified. If we make the assumption that useful features from the sequences are easily classifiable compared to background features, then a smaller value of $C_1$ could help find the decision hyperplane easily. However, the correct value of $C_1$ depends on each sequence. Thus, in Algorithm~\eqref{alg1}, we propose a heuristic scheme to find the SVMP descriptor for a given sequence $\pseq{}$ by iteratively tuning $C_1$ such that at least a fraction $\eta$ of the features in the positive bag are classified as positive. 

\begin{algorithm}  
	\SetAlgoLined
	\KwIn{$\pseq{}$, $\ndataset$, $\eta$}
	$C_1 \leftarrow \epsilon,\ \lambda > 1$\;
	\Repeat{$\frac{\card{\hpseq{}}}{\card{\pseq{}}}\geq \eta$} {
		$C_1 \leftarrow \lambda C_1$\;
		$[w,b] \leftarrow \argmin_{w,b} \svm(\pseq{},\ \ndataset,\ C_1)$\;
		$\hpseq{} \leftarrow \set{\feat\in\pseq{}\ |\ w^T\feat+b \geq 0}$\;
	}
	\KwRet{$[w,b]$}
	\caption{Efficient solution to the MIL problem $P1$}
	\label{alg1}
\end{algorithm}
 
Each step of Algorithm~\eqref{alg1} solves a standard SVM objective. Suppose we have an oracle that could give us a fixed value $C$ for $C_1$ that works for all action sequences for a fixed $\eta$. As is clear, there could be multiple combinations of data points in $\hpseq{}$ that could satisfy this $\eta$. If $\hpseq{p}$ is one such $\hpseq{}$. Then, $P1$ using $\hpseq{p}$ is just the SVM formulation and is thus convex. That is, if we enumerate all such $\hpseq{p}$ that satisfies the constraint using $\eta$, then the objective for each such $\hpseq{p}$ is an SVM problem, that could be solved using standard efficient solvers. Instead of enumerating all such bags $\hpseq{p}$, in Alg.~\ref{alg1}, we adjust the SVM classification rate to $\eta$, which is easier to implement. Assuming we find a $C_1$ that satisfies the $\eta$-constraint using $P1$, due to the convexity of SVM, it can be shown that the objective of P1 will be the same in both cases (exhaustive enumeration and our proposed regularization adjustment), albeit the solution $\hat{X}_p^+$ might differ (there could be multiple solutions). 

Considering $P2$, it is non-convex in $Z$ and $(w_i,b_i)$'s jointly. However, it is convex in $Z$ when fixing $(w_i,b_i),\forall i\in\set{1,2,\cdots, N}$. Thus, under the above conditions, if we need to run only one iteration of $P1$, then $P2$ becomes convex in either variables separately, and thus we could solve it using block coordinate descent (BCD) towards a local minimum. Algorithm~\ref{alg2} depicts the iterations. Note that there is a coupling between the data point decision boundaries $(w_i,b_i)$ and the action classifier decision boundaries $Z_j$ in~\eqref{eq:9}, either of which are fixed when optimizing over the other using BCD. When optimizing over $(w_i,b_i)$, $Z_j^T\wbstack_l$ (in~\eqref{eq:9}) is a constant, and we use $\Delta(y_i^+, y_i^+)=1$, in which case the problem is equivalent to assuming $Z$ as a~\emph{virtual} positive data point in the positive bag. We make use of this observation in Algorithm~\ref{alg2} by including $Z$ in the positive bag. Note that these virtual $Z$ points are updated in place rather than adding new points in every iteration.


\begin{algorithm}  
	\SetAlgoLined
	\KwIn{$\pdataset$, $\ndataset$, $\eta$}
	\Repeat{until convergence} {
		\tcc{compute SVMP descriptors for all sequences}
		\For{$\pseq{i}\in\pdataset$}
		{
			$[w_i,b_i] \leftarrow \argmin_{w,b} \svm(\pseq{i},\ \ndataset,\ C)$\;
		}
		$Z\ \leftarrow\ \text{Solve P2 fixing $\wbstack_i, \forall i=\set{1,\cdots, N}$}$\;
		\tcc{$Z$ is added to $\pseq{i}$ so that $\svm$ could be used to satify~\eqref{eq:9}}
		$\pseq{i}\leftarrow \pseq{i} \cup Z$ 	
	}
	\KwRet{$Z$}
	\caption{A block-coordinate scheme for $P2$}
	\label{alg2}
\end{algorithm}

When using decision boundaries as data descriptors, a natural question can be regarding the identifiability of the sequences using this descriptor, especially if the negative bag is randomly sampled. To circumvent this issue, we propose two workarounds, namely (i) to use the same negative bag for all the sequences, and (ii) assume all features (including positives and negatives) are centralized with respect to a global data mean.

\subsection{Nonlinear Extensions}

In problem $P1$, we assume a linear decision boundary generating SVMP descriptors. However, looking back at our solutions in Algorithms~\eqref{alg1} and~\eqref{alg2}, it is clear that we are dealing with standard SVM formulations to solve our relaxed objectives. In the light of this, instead of using linear hyperplanes for classification, we may use nonlinear decision boundaries by using the kernel trick to embed the data in a Hilbert space for better representation. Assuming $\dataset=\pdataset\cup\ndataset$, by the Representer theorem~\cite{smola1998learning}, it is well-known that for a kernel $K:\dataset\times \dataset\rightarrow \reals{}_+$, the decision function $f$ for the SVM problem P1 will be of the form:
\begin{equation}
f(.) = \sum_{\feat\in\pseq{}\cup\ndataset}\alpha_{\feat} K(., \feat),
\label{eq:ksvm}
\end{equation}
where $\alpha_{\feat}$ are the parameters of the non-linear decision boundaries. However, from an implementation perspective, such a direct kernelization may be problematic, as we will need to store the training set to construct the kernel. We avoid this issue by restricting our formulation to use only homogeneous kernels~\cite{vedaldi2012efficient}, as such kernels have explicit linear feature map embeddings on which a linear SVM can be trained directly. This leads to exactly the same formulations as in~\eqref{eq:mil}, except that now our features $\feat$ are obtained via a homogeneous kernel map. In the sequel, we call such a descriptor a~\emph{nonlinear SVM pooling} (NSVMP) descriptor. 

%% file: end2end.tex
\section{End-to-End CNN Learning}
\label{e2e}
In this section, we address the problem of training a CNN end-to-end with SVM pooling as an intermediate layer -- the main challenge is to derive the gradients of SVMP for efficient backpropagation. Assume a CNN $f$ taking a sequence $S$ as input. Let $f_{L}$ denote the $L$-th CNN layer and let $Y_{L}$ denote the feature maps generated by this layer for all frames in $S$. We assume these features go into an SVMP layer and produces as output a descriptor $w$ (using a precomputed set of negative feature maps), which is then passed to subsequent CNN layers for classification. Mathematically, let 
$g(w) = \argmin_{w} \svmp(Y_{L})$ define the SVM pooling layer, which we re-define the hinge-loss as: 
\begin{equation}
\svmp(Y_{L})=\frac{1}{2}\enorm{w}^2+ \frac{\lambda}{2}\sum_{z\in Y_{L}}\!\!\max\left(0,\theta(z;\eta)w^Tz-1\right)^2.\nonumber
\end{equation}


As is by now clear, with regard to a CNN learning setup, we are dealing with a bilevel optimization problem here -- that is, optimizing for the CNN parameters via stochastic gradient descent in the outer optimization, which requires the gradient of an argmin inner optimization with respect to its optimum, i.e., we need to compute the gradient of $g(w)$ with respect to the data $z$. By applying Lemma 3.3 of~\cite{gould2016differentiating}, this gradient of the argmin at an optimum SVMP solution $w^*$ can be shown to be the following:
\begin{equation}
\nabla_{z} g(w^*) = -\nabla_{ww} \svmp(Y_{L})^{\!\!-1} \nabla_{zw} \svmp(Y_{L}),\nonumber
\end{equation}
where the first term captures the inverse of the Hessian evaluated at $w^*$ and the second term is the second-order derivative wrt $z$ and $w$. Substituting for the components, we have the gradient at $w=w^*$ as:
{\small 
\begin{align}
-\!\!\left(\!\!\eye{}\!\!+\!\!\lambda\hspace*{-0.5cm}\sum_{\forall j:\theta_jw^Tz_j >1}\hspace*{-0.5cm} (\theta_jz_j)(\theta_jz_j)^T\right)^{\!\!\!\!-1}\!\!\!\left[\lambda\hspace*{-0.3cm}\sum_{\forall j: \theta_jw^Tz_j >1} \hspace*{-0.6cm}\text{D }(\theta_j^2w^Tz_j\!-\!\theta_j)\!+\!\theta_j^2wz_j^T\!\!\right]\notag
\label{eq:bilevel}
\end{align}
}
\noindent where for brevity, we use $\theta_j = \theta(z_j; \eta)$, and $\text{D}$ is a diagonal matrix, whose $i$-th entry as $D_{ii}=\theta_i^2w^Tz_i-\theta_i$. 

%% file: expts.tex
\section{Experiments}
\label{sec:exp}
In this section, we explore the utility of discriminative pooling on several tasks, namely (i) action recognition using video and skeletal features, (ii) action anticipation, and (iii) localizing actions in videos. We introduce these datasets briefly next along with details of the features used, followed by an analysis of the parameters of our pooling scheme, before furnishing our results against state-of-the-art. 

\subsection{Datasets}
\label{dataset}

\noindent\textbf{HMDB-51~\cite{kuehne2011hmdb}:} is a popular benchmark for video action recognition, consisting of trimmed videos downloaded from the Internet. The dataset contains 51 action classes and 6766 videos. The recognition results are evaluated using 3-fold cross-validation and mean classification accuracy is reported. For this dataset, we analyze different combinations of features on multiple CNN frameworks.

\noindent\textbf{Charades~\cite{sigurdsson2016hollywood}:} is an untrimmed multi-action dataset, containing 11,848 videos split into 7985 for training, 1863 for validation, and 2,000 for testing. It has 157 action categories, including several fine-grained classes. In the classification task, we follow the evaluation protocol of ~\cite{sigurdsson2016hollywood}, using the output probability of the classifier to be the score of the sequence. In the detection task, we use 'post-processing' protocol described in~\cite{Sigurdsson_2017_CVPR}, which uses the averaged prediction score of a small temporal window around temporal pivots. The dataset provides two-stream VGG-16 fc7 features which we use in our method.\footnote{http://vuchallenge.org/charades.html} The performance on detection and recognition tasks are evaluated using mean average precision (mAP) on the validation set. 

\noindent\textbf{NTU-RGBD~\cite{shahroudy2016ntu}:} is by far the largest action datasets providing 3D skeleton data. It has 56,000 videos and 60 actions performed by 40 people from 80 different views. We use the temporal CNN proposed in~\cite{kim2017interpretable} to generate features, but uses SVMP instead of their global average pooling.

\begin{figure}[htbp]
	\begin{center}
        \subfigure[]{\label{subfig:1}\includegraphics[width=4.1cm,trim={1cm 0cm 3.3cm 0cm},clip]{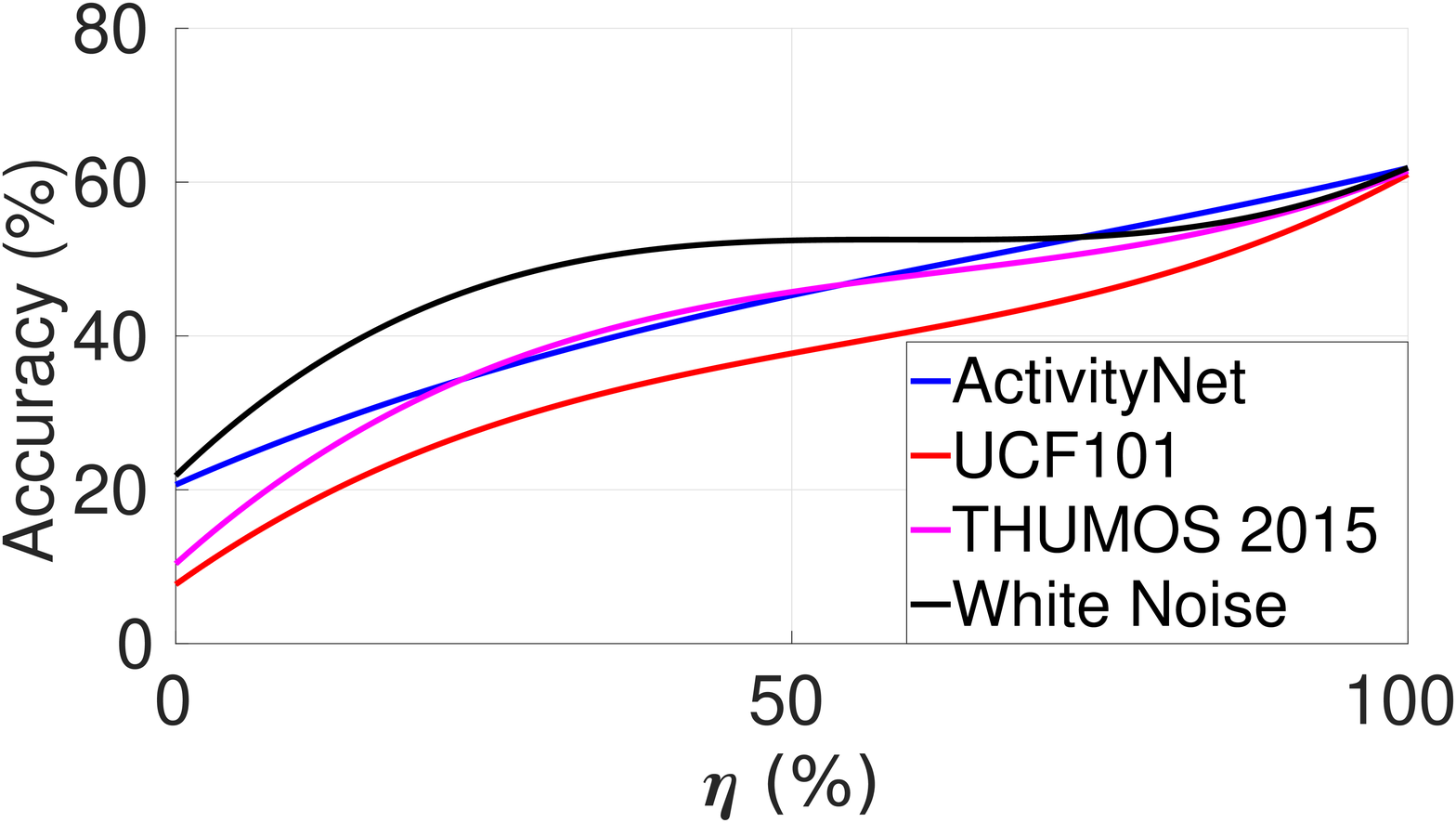}}
        \subfigure[]{\label{subfig:2}\includegraphics[width=4.1cm,trim={1cm 0cm 4cm 0cm},clip]{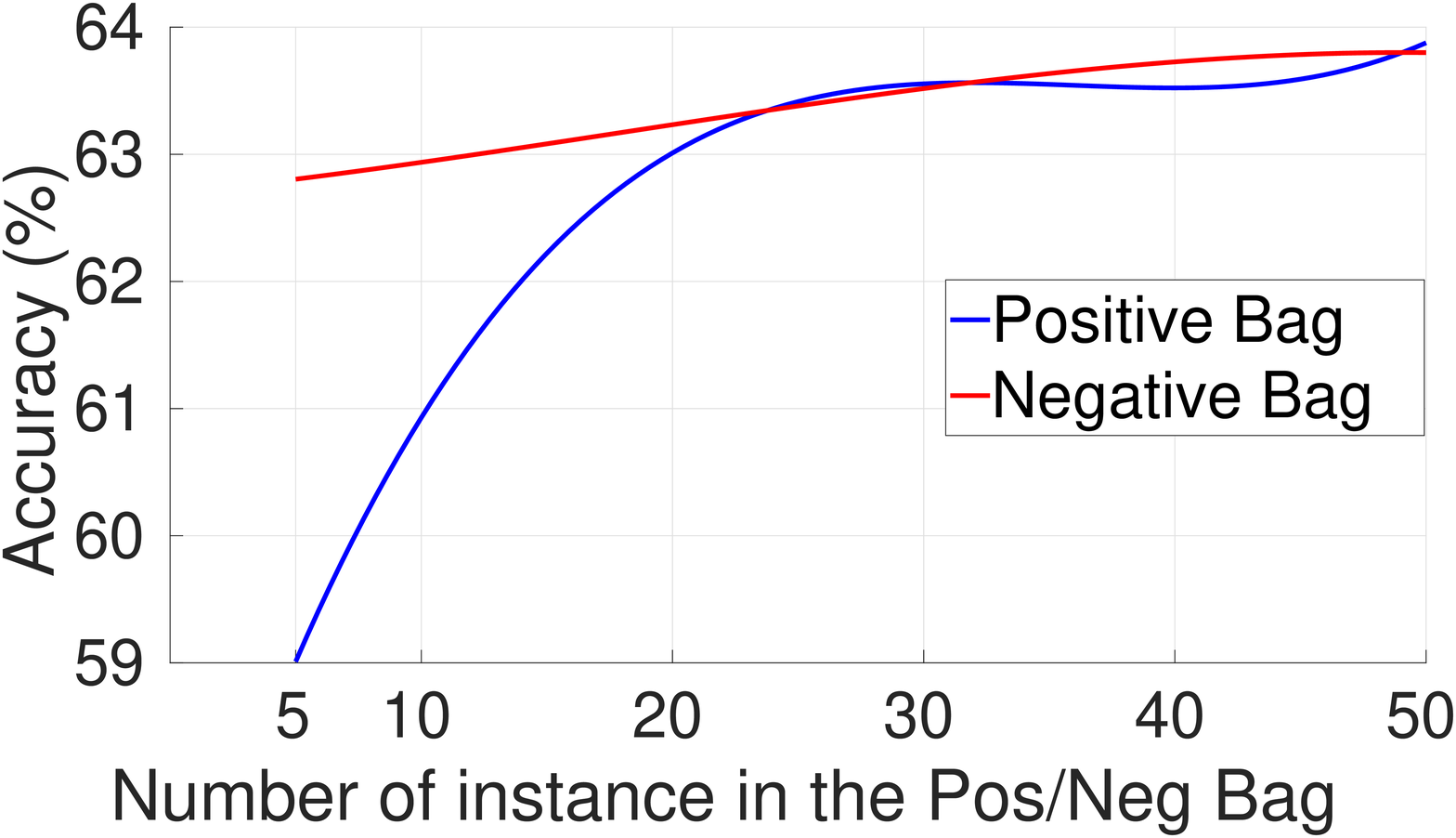}}
     \subfigure[]{\label{subfig:3}\includegraphics[width=4.1cm,trim={1cm 0cm 4cm 0cm},clip]{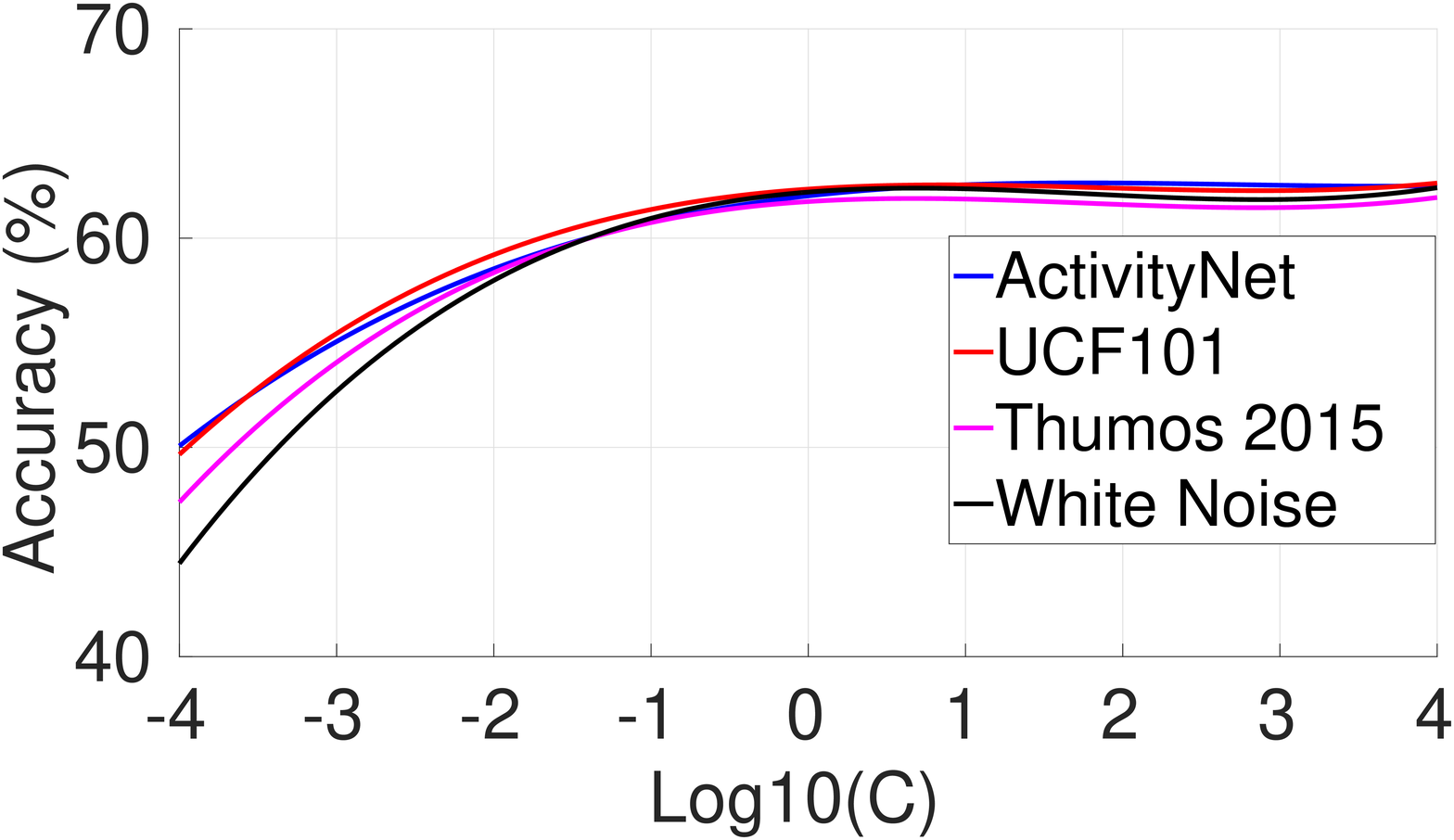}}
    \subfigure[]{\label{subfig:4}\includegraphics[width=4.1cm,trim={1cm 0cm 2.8cm 0cm},clip]{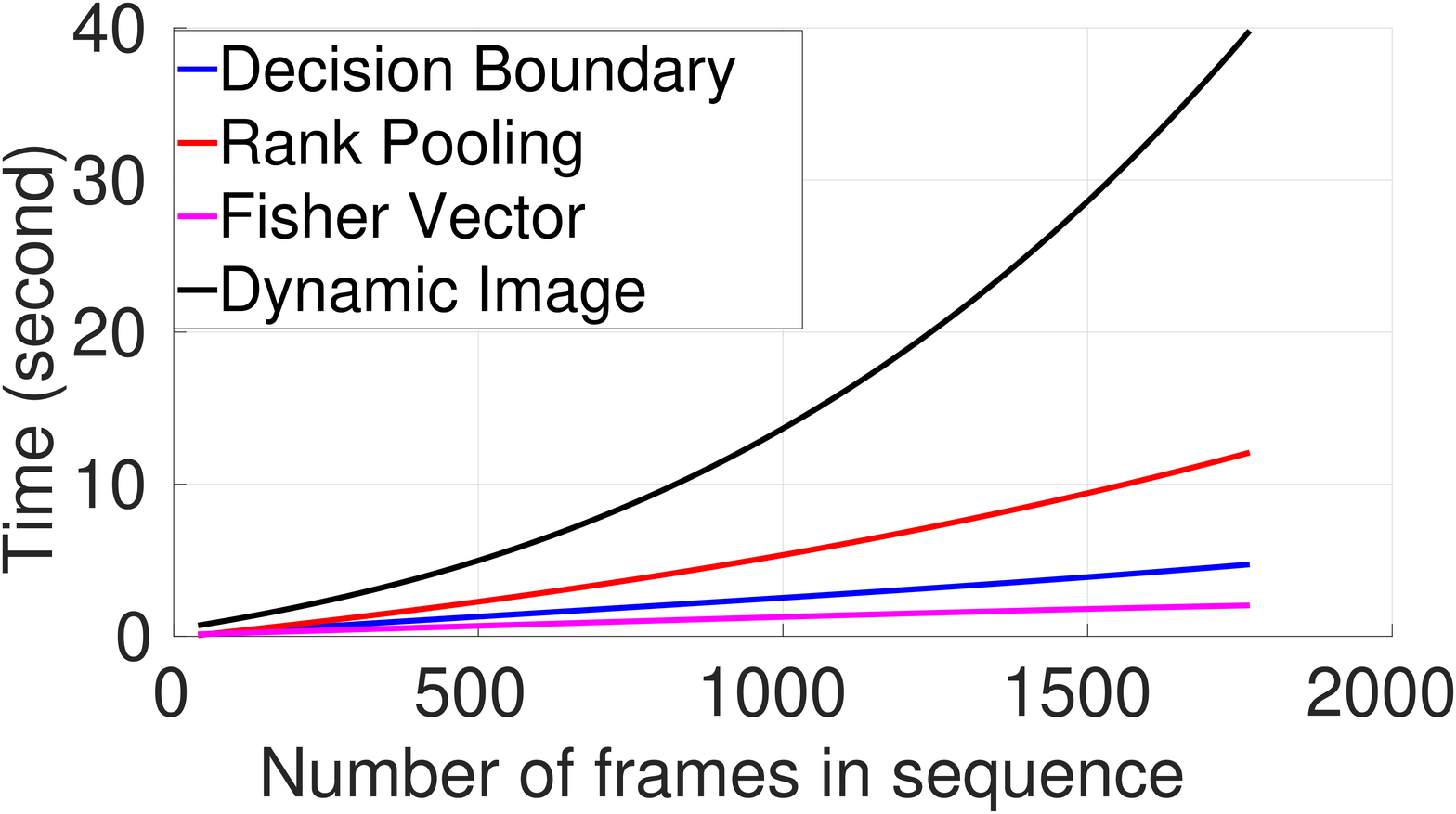}}        
	\end{center}
	\caption{Analysis of the parameters in our scheme. All experiments use VGG features from fc6 layer. See text for details.}
	\label{fig:all_plots}
\end{figure}

\subsection{Parameter Analysis}
In this section, we analyze the influence of each of the parameters in our scheme.

\noindent\textbf{Selecting Negative Bags:} An important step in our algorithm is the selection of the positive and negative bags in the MIL problem. We randomly sample the required number of frames (50) from each sequence/fold in the training/testing set to define the positive bags. In terms of the negative bags, we need to select samples that are unrelated to the ones in the positive bags. We explored four different negatives in this regard to understand the impact of this selection and apply them on the HMDB-51 dataset split-1. They are samples from (i) ActivityNet dataset~\cite{caba2015activitynet} unrelated to HMDB-51, (ii) UCF-101 dataset unrelated to HMDB-51, (iii) Thumos Challenge background sequence\footnote{http://www.thumos.info/home.html}, and (iv) synthesized random white noise image sequences. For (i) and (ii), we use 50 frames each from randomly selected videos, one from every unrelated class, and for (iv) we used 50 synthesized white noise images, and randomly generated stack of optical flow images. As shown in Figure ~\ref{subfig:1}, the white noise negative shows better performance for both lower and higher value of $\eta$ parameter. So, we use it in our experiments for other datasets. 

\noindent\textbf{Choosing Hyperparameters:} The three important parameters in our scheme are (i) the $\eta$ deciding the quality of an SVMP descriptor, (ii) $C_1=C$ used in Algorithm~\ref{alg1} when finding SVMP per sequence, and (iii) sizes of the positive and negative bags. To study (i) and (ii), we plot in Figures~\ref{subfig:3} and~\ref{subfig:1} for HMDB-51 dataset, classification accuracy when $C$ is increased from $10^{-4}$ to $10^{4}$ in steps and when $\eta$ is increased from 0-100\% and respectively. We repeat this experiment for all the different choices of negative bags. As is clear, increasing these parameters reduces the training error, but may lead to overfitting. However, Figure~\ref{subfig:2} shows that increasing $C$ increases the accuracy of the SVMP descriptor, implying that the CNN features are already equipped with discriminative properties for action recognition. However, beyond $C=10$, a gradual decrease in performance is witnessed, suggesting overfitting to bad features in the positive bag. Thus, we use $C=10$ ( and $\eta=0.9$) in the experiments to follow. To decide the bag sizes for MIL, we plot in Figure~\ref{subfig:2}, performance against increasing size of the positive bag, while keeping the negative bag size at 50 and vice versa; i.e., for the red line in Figure \ref{subfig:2}, we fix the number of instances in the positive bag at 50; we see that the accuracy raises with the cardinality of the negative bag. A similar trend, albeit less prominent is seen when we repeat the experiment with the negative bag size, suggesting that about 30 frames per bag is sufficient to get a useful descriptor. 

\noindent\textbf{Running Time:} In Figure~\ref{subfig:4}, we compare the time it took on average to generate SVMP descriptors for an increasing number of frames in a sequence. For comparison, we plot the running times for some of the recent pooling schemes such as rank pooling~\cite{bilen2016dynamic,fernando2015modeling} and the Fisher vectors~\cite{wang2013action}. The plot shows that while our scheme is slightly more expensive than standard Fisher vectors (using the VLFeat\footnote{http://www.vlfeat.org/}), it is significantly cheaper to generate SVMP descriptors in contrast to some of the recent popular pooling methods. 

\subsection{Experiments on HMDB-51}
Following the recent trends, for this experiment, we use a two-stream CNN model in popular architectures, the VGG-16 and the ResNet-152~\cite{feichtenhofer2016convolutional,simonyan2014very}. We fine-tune a two-stream VGG/ResNet model trained for the UCF-101 dataset. 

\noindent\textbf{SVMP on Different CNN Features:} We generate SVMP descriptors from different intermediate layers of the CNN models and compare their performance. Specifically, features from each layer are used as the positive bags and SVMP descriptors computed using Algorithm~\ref{alg1} and~\ref{alg2} against the chosen set of negative bags. In Table~\ref{table:1}, we report results on split-1 of the HMDB-51 dataset and find that the combination of fc6 and pool5 gives the best performance for the VGG-16 model, while pool5 features alone show good performance using ResNet. We thus use these feature combinations for experiments to follow. 

\begin{table}[]
\centering
\caption{Comparisons using various features on HMDB-51 split-1}
\label{table:1}\scalebox{0.85}{
\begin{tabular}{lcc}
\hline
Feature/   & Accuracy   & Accuracy when\\
model        & independently & combined with:\\
\hline
pool5 (vgg-16)   & 57.9\%         & \textbf{63.8\%} (fc6)                                  \\
fc6 (vgg-16)     & 63.3\%      & -                              \\
fc7 (vgg-16)     & 56.1\%         & 57.1\% (fc6)                                 \\
fc8 (vgg-16)     & 52.4\%         & 58.6\% (fc6)                                 \\
softmax (vgg-16) & 41.0\%         & 46.2\% (fc6)   \\
\hline
pool5 (ResNet-152)  & \textbf{69.5\%}         & -                               \\
fc1000 (ResNet-152)    & 61.1\%    & 68.8\% (pool5)
\end{tabular}}
\end{table}


\noindent\textbf{SVMP Extensions and Standard Pooling:} We analyze the complementary nature of SVMP and its non-linear extension NSVMP (using a Chi-sq homogeneous kernel) on HMDB-51 split1. The results are provided in Table~\ref{table:3}, and clearly show that the combination leads to significant improvements consistently on both datasets. Comparison between SVMP and standard pooling schemes (such as average (AP) and max (MP)) are reported in Table~\ref{table:4} using exactly the same set of features. As is clear, SVMP is significantly better than the other two pooling schemes.
\begin{table}[]
\centering
\caption{Comparisons between SVMP and NSVMP on HMDB-51 split-1}
\label{table:3}\scalebox{0.85}{
\begin{tabular}{l|ll}
\hline
       & VGG & ResNet  \\\hline
Linear-SVMP   & 63.8\% & 69.5\%  \\
Non-linear-SVMP  & 64.4\% & 69.8\%  \\
Combination & \textbf{66.1\%}&  \textbf{71.0\%}     
\end{tabular}}
\end{table}


\begin{table}[]
\centering
\caption{Comparison to standard pooling on HMDB-51 split-1}
\label{table:4}\scalebox{0.85}{
\begin{tabular}{l|ll}
\hline
       & VGG & ResNet  \\\hline
Spatial Stream-AP\cite{feichtenhofer2016spatiotemporal,feichtenhofer2016convolutional}   
	   & 47.1\% & 46.7\%  \\
Spatial Stream-MP  
	   & 46.5\% & 45.1\%  \\
Spatial Stream-SVMP  
	   & \textbf{58.3\%} & \textbf{57.4\%}  \\
       \hline
Temporal Stream-AP \cite{feichtenhofer2016spatiotemporal,feichtenhofer2016convolutional} 
       & 55.2\% & 60.0\% \\
Temporal Stream-MP
       & 54.8\% & 58.5\%  \\
Temporal Stream-SVMP
       & \textbf{61.8\%} &\textbf{65.7\%} \\
       \hline
Two-Stream-AP \cite{feichtenhofer2016spatiotemporal,feichtenhofer2016convolutional} 
       & 58.2\%& 63.8\% \\
Two-Stream-MP
       & 56.7\%& 60.6\%  \\
Two-Stream-SVMP
       &\textbf{66.1\%}&\textbf{71.0\%}
\end{tabular}}
\end{table}


\noindent\textbf{SVMP for Action Anticipation}
We also evaluated the usefulness of SVMP for action anticipation. This is motivated by the intuition that SVMP might be able to learn generalizable decision boundaries when shown only a small part of the sequence -- given the SVM is optimized in a max-margin framework. Specifically, we use $k\times\frac{1}{5}$ initial part of the sequences to be pooled by SVMP, ($k\in\set{1,2,3,4,5}$) which has to now predict the action in the full segment.  We use the ResNet feature for this experiment. The results are provided in Table~\ref{tab:4.5} and is clear that compared with others, the benefits of SVMP become higher, when only seeing a small fraction of the data, substantiating our intuition. 


\begin{table}[]
\centering
\caption{Comparison of action anticipation on HMDB-51 split-1}
\label{tab:4.5}\scalebox{0.85}{
\begin{tabular}{l|lllll}\hline
\multicolumn{6}{c}{HMDB-51}                           \\\hline
k/5 & 1/5    & 2/5    & 3/5    & 4/5    & 1      \\\hline
SVMP     & \textbf{58.3\%} & \textbf{65.5\%} & \textbf{68.4\%} & \textbf{70.1\%} & \textbf{71.0\%} \\
AP       & 48.6\% & 56.4\% & 59.9\% & 62.5\% & 63.8\% \\
MP       & 46.2\% & 55.4\% & 56.3\% & 58.8\% & 60.6\% \\
\end{tabular}}
\end{table}
\subsection{Recognition/Detection in Untrimmed Videos} As introduced in the Section~\ref{dataset}, Charades is an untrimmed dataset with multiple actions in one sequence. We use the publicly available two-stream VGG features from the fc7 layer for this dataset. We applied our scheme on the provided training set (7985 videos), and report results (mAP) on the provided validation set (1863 videos) for the tasks of action classification and detection. In the classification task, we concatenate the two-stream features and apply a sliding pooling scheme to create multiple descriptors. Following the evaluation protocol in~\cite{sigurdsson2016hollywood}, we use the output probability of the classifier to be the score of the sequence. In the detection task, the standard evaluation setting is to use the prediction score of 25 equidistant time points in the sequence, which is not suitable for any pooling scheme. So, we consider another evaluation method with post-processing, proposed in~\cite{Sigurdsson_2017_CVPR}. This method uses the averaged prediction score of a small temporal window around each temporal pivots. Instead of average pooling, we apply the SVMP. From Table~\ref{table:10}, it is clear that SVMP improves performance against other pooling schemes.
\subsection{Skeletal Action Recognition in NTU-RGBD}
For this experiment, we follow the two official evaluation protocols described in~\cite{shahroudy2016ntu}, i.e., the cross-view and cross-subject protocol. We use~\cite{kim2017interpretable} as the baseline. This scheme applies a temporal CNN with residual connections on the 3D skeleton data. We swap the global average pooling layer in~\cite{kim2017interpretable} by a Rank/SVM pooling layer. The result in Table~\ref{table:10} indicates that the SVMP works better than other pooling schemes on the skeleton-based features. 

\subsection{Visualization of SVMP}
\label{discussion}

\begin{figure}[]
	\begin{center}
        \includegraphics[width=1\linewidth,trim={0cm 0cm 0cm 0cm},clip]{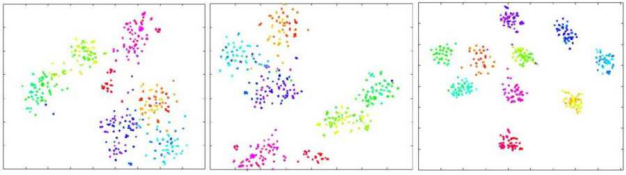}
	\end{center}
	\caption{T-SNE visualizations of SVMP and other pooling methods. From left to right: average pooling, max pooling, and SVMP.}
   	\label{fig:4}
\end{figure}

To gain further intuitions into the performance boost by SVMP, 
in Figure~\ref{fig:4}, we show TSNE visualizations comparing to average and max pooling on 10-classes from HDMB-51. The visualization shows that SVMP leads to better separated clusters, substantiating that it is learning much more discriminative representations than traditional methods.
\subsection{Comparisons to the State of the Art}
In Table \ref{table:10}, we compare our best result against the state-of-the-art results on each dataset using the respective standard evaluation protocols. For a fair comparison, we also report our best result combining with hand-crafted features (IDT-FV)~\cite{wang2013dense} for HMDB-51. Our scheme obtains the state-of-the-art performance in all datasets and outperform other methods by 1--4\%. We note that recently the two-stream I3D+ model\cite{carreira2017quo}, which is pre-trained on the larger Kinectics dataset (with more than 300K videos), achieves 80\% on HMDB-51. However, without additional data, two-stream I3D is outperformed by our SVMP. Moreover, most of these methods could enjoy a further boost by applying our SVMP scheme. To substantiate this, we also show the I3D+ model to use SVMP (instead of their proposed average pooling) on HMDB-51 dataset using the settings in~\cite{carreira2017quo}. 


\begin{table}[]
\centering
\caption{Comparison to the state of the art in each dataset, following the official evaluation protocol for each dataset.}
\label{table:10}
\scalebox{0.85}{
\begin{tabular}{@{}l|c|c@{}}\hline
\multicolumn{3}{c}{HMDB-51 (accuracy over 3 splits)}                      \\\hline
Method                       & \multicolumn{2}{c}{Accuracy}         \\\hline
Temporal segment networks\cite{Wang2016}    &  \multicolumn{2}{c}{69.4\%}           \\
AdaScan\cite{Kar_2017_CVPR}           &  \multicolumn{2}{c}{54.9\%}             \\
AdaScan + IDT + C3D\cite{Kar_2017_CVPR}           & \multicolumn{2}{c}{ 66.9\% }            \\
ST ResNet\cite{feichtenhofer2016spatiotemporal}                   & \multicolumn{2}{c}{66.4\%}              \\
ST ResNet + IDT\cite{feichtenhofer2016spatiotemporal}                   & \multicolumn{2}{c}{ 70.3\%}             \\
ST Multiplier Network\cite{feichtenhofer2017spatiotemporal}        & \multicolumn{2}{c}{ 68.9\% }              \\
ST Multiplier Network + IDT\cite{feichtenhofer2017spatiotemporal}        & \multicolumn{2}{c}{ 72.2\% }          \\
Two-stream I3D\cite{carreira2017quo}               & \multicolumn{2}{c}{ 66.4\% }          \\
Two-stream I3D+ (Kinetics 300k)\cite{carreira2017quo}               & \multicolumn{2}{c}{80.9\%} \\\hline
SVMP (ResNet)                  & \multicolumn{2}{c}{71.0\%}            \\
SVMP (ResNet+IDT)              & \multicolumn{2}{c}{ \textbf{72.6}\% }             \\
SVMP (I3D+)              & \multicolumn{2}{c}{ \textbf{81.3}\% }             \\\hline
\multicolumn{3}{c}{Charades (mAP)}                                \\\hline
Method                       & Classification         & Detection        \\\hline
Two-stream VGG (Average Pooling)~\cite{simonyan2013deep}   & 14.3\%       & 10.9\%       \\
Two-stream VGG (Max Pooling)~\cite{simonyan2013deep}   & 15.3\%       & 9.2\%       \\
ActionVLAD + IDT\cite{girdhar2017actionvlad}             & 21.0\%           & -    \\
Asynchronous Temporal Fields~\cite{Sigurdsson_2017_CVPR} & 22.4\%           & 12.8\%       \\\hline
SVMP(VGG)                  & 25.1\%           & 13.9\%    \\
SVMP(VGG+IDT)              & \textbf{26.7\%}  & \textbf{14.2\%} \\\hline
\multicolumn{3}{c}{NTU-RGBD}                                \\\hline
Method                       & Cross-Subject  & Cross-View  \\\hline
Res-TCN (Average Pooling)\cite{kim2017interpretable}                 & 74.3\%         & 83.1\%      \\
Res-TCN (Rank Pooling~\cite{bilen2016dynamic})                  &75.5\%         & 83.9\%      \\
STA-LSTM~\cite{song2017end} &73.4\% &81.2\% \\
ST-LSTM + Trust Gate\cite{liu2017skeleton}    & 69.2\%         & 77.7\%      \\
Body-parts learning~\cite{rahmani2017learning} & 75.2\% &83.1\% \\\hline
SVMP (Res-TCN)                  & \textbf{78.5\%}         & \textbf{86.4\%}      \\\hline
\end{tabular}}
\end{table}

%% file: conclude.tex
\section{Conclusion}
\label{sec:conclude}
In this paper, we presented a simple, efficient, and powerful pooling scheme, SVM pooling, for summarizing videos. We cast the pooling problem in a multiple instance learning framework, and seek to learn useful decision boundaries on the frame level features from each sequence against background/noise features. We provide an efficient scheme that jointly learns these decision boundaries and the action classifiers on them. We also extended the framework to deal with nonlinear decision boundaries and  end-to-end CNN training. Extensive experiments were showcased on three challenging benchmark datasets, demonstrating state-of-the-art performance. Given the challenging nature of these datasets, we believe the benefits afforded by our scheme is a significant step towards the advancement of recognition systems designed to represent videos.

\comment{
It summarize actions in video sequences by filtering useful features from a bag of per-frame CNN features. And We propose to use the output of SVM pooling as a descriptor for representing the sequence, namely the SVM Pooled (SVMP) descriptor. Considering SVMP descriptor as a better representation than CNN features, we extended SVMP to NSVMP to deal with the  non-linear sequences. Extensive experiments were implemented on two popular benchmark datasets: HMDB51 and UCF101, which clearly show the advantage and effectiveness of SVMP compared with the traditional pooling scheme on CNN features. More importantly, after combining with the hand-crafted local features, we reach the state-of-the-art result on these two datasets.
}

%% file: dbcnn_cvpr18.bbl
\begin{thebibliography}{10}\itemsep=-1pt

\bibitem{baccouche2011sequential}
M.~Baccouche, F.~Mamalet, C.~Wolf, C.~Garcia, and A.~Baskurt.
\newblock Sequential deep learning for human action recognition.
\newblock In {\em Human Behavior Understanding}, pages 29--39. 2011.

\bibitem{bilen2016dynamic}
H.~Bilen, B.~Fernando, E.~Gavves, A.~Vedaldi, and S.~Gould.
\newblock Dynamic image networks for action recognition.
\newblock In {\em CVPR}, 2016.

\bibitem{bunescu2007multiple}
R.~C. Bunescu and R.~J. Mooney.
\newblock Multiple instance learning for sparse positive bags.
\newblock In {\em ICML}, 2007.

\bibitem{caba2015activitynet}
F.~Caba~Heilbron, V.~Escorcia, B.~Ghanem, and J.~Carlos~Niebles.
\newblock Activitynet: A large-scale video benchmark for human activity
  understanding.
\newblock In {\em CVPR}, 2015.

\bibitem{carreira2017quo}
J.~Carreira and A.~Zisserman.
\newblock Quo vadis, action recognition? a new model and the kinetics dataset.
\newblock In {\em CVPR}, July 2017.

\bibitem{grp}
A.~Cherian, B.~Fernando, M.~Harandi, and S.~Gould.
\newblock Generalized rank pooling for activity recognition.
\newblock In {\em CVPR}, 2017.

\bibitem{cinbis2017weakly}
R.~G. Cinbis, J.~Verbeek, and C.~Schmid.
\newblock Weakly supervised object localization with multi-fold multiple
  instance learning.
\newblock {\em PAMI}, 39(1):189--203, 2017.

\bibitem{donahue2015long}
J.~Donahue, L.~Anne~Hendricks, S.~Guadarrama, M.~Rohrbach, S.~Venugopalan,
  K.~Saenko, and T.~Darrell.
\newblock Long-term recurrent convolutional networks for visual recognition and
  description.
\newblock In {\em CVPR}, 2015.

\bibitem{du2015hierarchical}
Y.~Du, W.~Wang, and L.~Wang.
\newblock Hierarchical recurrent neural network for skeleton based action
  recognition.
\newblock In {\em CVPR}, 2015.

\bibitem{feichtenhofer2016spatiotemporal}
C.~Feichtenhofer, A.~Pinz, and R.~Wildes.
\newblock Spatiotemporal residual networks for video action recognition.
\newblock In {\em NIPS}, 2016.

\bibitem{feichtenhofer2017spatiotemporal}
C.~Feichtenhofer, A.~Pinz, and R.~P. Wildes.
\newblock Spatiotemporal multiplier networks for video action recognition.
\newblock In {\em CVPR}, 2017.

\bibitem{feichtenhofer2017temporal}
C.~Feichtenhofer, A.~Pinz, and R.~P. Wildes.
\newblock Temporal residual networks for dynamic scene recognition.
\newblock In {\em CVPR}, 2017.

\bibitem{feichtenhofer2016convolutional}
C.~Feichtenhofer, A.~Pinz, and A.~Zisserman.
\newblock Convolutional two-stream network fusion for video action recognition.
\newblock In {\em CVPR}, 2016.

\bibitem{fernando2015modeling}
B.~Fernando, E.~Gavves, J.~M. Oramas, A.~Ghodrati, and T.~Tuytelaars.
\newblock Modeling video evolution for action recognition.
\newblock In {\em CVPR}, 2015.

\bibitem{gartner2002multi}
T.~G{\"a}rtner, P.~A. Flach, A.~Kowalczyk, and A.~J. Smola.
\newblock Multi-instance kernels.
\newblock In {\em ICML}, 2002.

\bibitem{girdhar2017actionvlad}
R.~Girdhar, D.~Ramanan, A.~Gupta, J.~Sivic, and B.~Russell.
\newblock Actionvlad: Learning spatio-temporal aggregation for action
  classification.
\newblock In {\em CVPR}, 2017.

\bibitem{gould2016differentiating}
S.~Gould, B.~Fernando, A.~Cherian, P.~Anderson, R.~S. Cruz, and E.~Guo.
\newblock On differentiating parameterized argmin and argmax problems with
  application to bi-level optimization.
\newblock {\em arXiv preprint arXiv:1607.05447}, 2016.

\bibitem{hayat2015deep}
M.~Hayat, M.~Bennamoun, and S.~An.
\newblock Deep reconstruction models for image set classification.
\newblock {\em PAMI}, 37(4):713--727, 2015.

\bibitem{Kar_2017_CVPR}
A.~Kar, N.~Rai, K.~Sikka, and G.~Sharma.
\newblock Adascan: Adaptive scan pooling in deep convolutional neural networks
  for human action recognition in videos.
\newblock In {\em CVPR}, 2017.

\bibitem{kim2017interpretable}
T.~S. Kim and A.~Reiter.
\newblock Interpretable 3d human action analysis with temporal convolutional
  networks.
\newblock {\em arXiv preprint arXiv:1704.04516}, 2017.

\bibitem{krizhevsky2012imagenet}
A.~Krizhevsky, I.~Sutskever, and G.~E. Hinton.
\newblock Imagenet classification with deep convolutional neural networks.
\newblock In {\em NIPS}, 2012.

\bibitem{kuehne2011hmdb}
H.~Kuehne, H.~Jhuang, E.~Garrote, T.~Poggio, and T.~Serre.
\newblock Hmdb: a large video database for human motion recognition.
\newblock In {\em ICCV}, 2011.

\bibitem{lai2014video}
K.-T. Lai, F.~X. Yu, M.-S. Chen, and S.-F. Chang.
\newblock Video event detection by inferring temporal instance labels.
\newblock In {\em CVPR}, 2014.

\bibitem{lazimy1982mixed}
R.~Lazimy.
\newblock Mixed-integer quadratic programming.
\newblock {\em Mathematical Programming}, 22(1):332--349, 1982.

\bibitem{li2016action}
Q.~Li, Z.~Qiu, T.~Yao, T.~Mei, Y.~Rui, and J.~Luo.
\newblock Action recognition by learning deep multi-granular spatio-temporal
  video representation.
\newblock In {\em ICMR}, 2016.

\bibitem{li2015multiple}
W.~Li and N.~Vasconcelos.
\newblock Multiple instance learning for soft bags via top instances.
\newblock In {\em CVPR}, 2015.

\bibitem{li2013dynamic}
W.~Li, Q.~Yu, A.~Divakaran, and N.~Vasconcelos.
\newblock Dynamic pooling for complex event recognition.
\newblock In {\em ICCV}, 2013.

\bibitem{liu2017skeleton}
J.~Liu, A.~Shahroudy, D.~Xu, A.~C. Kot, and G.~Wang.
\newblock Skeleton-based action recognition using spatio-temporal lstm network
  with trust gates.
\newblock {\em arXiv preprint arXiv:1706.08276}, 2017.

\bibitem{malisiewicz2011ensemble}
T.~Malisiewicz, A.~Gupta, and A.~A. Efros.
\newblock Ensemble of exemplar-svms for object detection and beyond.
\newblock In {\em ICCV}, 2011.

\bibitem{misener2013glomiqo}
R.~Misener and C.~A. Floudas.
\newblock Glomiqo: Global mixed-integer quadratic optimizer.
\newblock {\em Journal of Global Optimization}, 57(1):3--50, 2013.

\bibitem{nowozin2007discriminative}
S.~Nowozin, G.~Bakir, and K.~Tsuda.
\newblock Discriminative subsequence mining for action classification.
\newblock In {\em ICCV}, 2007.

\bibitem{pascanu2013difficulty}
R.~Pascanu, T.~Mikolov, and Y.~Bengio.
\newblock On the difficulty of training recurrent neural networks.
\newblock In {\em ICML}, 2013.

\bibitem{rahmani2017learning}
H.~Rahmani and M.~Bennamoun.
\newblock Learning action recognition model from depth and skeleton videos.
\newblock In {\em ICCV}, 2017.

\bibitem{sadanand2012action}
S.~Sadanand and J.~J. Corso.
\newblock Action bank: A high-level representation of activity in video.
\newblock In {\em CVPR}, 2012.

\bibitem{satkin2010modeling}
S.~Satkin and M.~Hebert.
\newblock Modeling the temporal extent of actions.
\newblock In {\em ECCV}, 2010.

\bibitem{schindler2008action}
K.~Schindler and L.~Van~Gool.
\newblock Action snippets: How many frames does human action recognition
  require?
\newblock In {\em CVPR}, 2008.

\bibitem{shahroudy2016ntu}
A.~Shahroudy, J.~Liu, T.-T. Ng, and G.~Wang.
\newblock Ntu rgb+ d: A large scale dataset for 3d human activity analysis.
\newblock In {\em CVPR}, 2016.

\bibitem{Sigurdsson_2017_CVPR}
G.~A. Sigurdsson, S.~Divvala, A.~Farhadi, and A.~Gupta.
\newblock Asynchronous temporal fields for action recognition.
\newblock In {\em CVPR}, 2017.

\bibitem{sigurdsson2016hollywood}
G.~A. Sigurdsson, G.~Varol, X.~Wang, A.~Farhadi, I.~Laptev, and A.~Gupta.
\newblock Hollywood in homes: Crowdsourcing data collection for activity
  understanding.
\newblock In {\em ECCV}, 2016.

\bibitem{simonyan2013deep}
K.~Simonyan, A.~Vedaldi, and A.~Zisserman.
\newblock Deep inside convolutional networks: Visualising image classification
  models and saliency maps.
\newblock {\em arXiv preprint arXiv:1312.6034}, 2013.

\bibitem{simonyan2014two}
K.~Simonyan and A.~Zisserman.
\newblock Two-stream convolutional networks for action recognition in videos.
\newblock In {\em NIPS}, 2014.

\bibitem{simonyan2014very}
K.~Simonyan and A.~Zisserman.
\newblock Very deep convolutional networks for large-scale image recognition.
\newblock {\em arXiv preprint arXiv:1409.1556}, 2014.

\bibitem{smola1998learning}
A.~J. Smola and B.~Sch{\"o}lkopf.
\newblock {\em Learning with kernels}.
\newblock Citeseer, 1998.

\bibitem{song2017end}
S.~Song, C.~Lan, J.~Xing, W.~Zeng, and J.~Liu.
\newblock An end-to-end spatio-temporal attention model for human action
  recognition from skeleton data.
\newblock In {\em AAAI}, 2017.

\bibitem{srivastava2015unsupervised}
N.~Srivastava, E.~Mansimov, and R.~Salakhutdinov.
\newblock Unsupervised learning of video representations using lstms.
\newblock In {\em ICML}, pages 843--852, 2015.

\bibitem{sun2014discover}
C.~Sun and R.~Nevatia.
\newblock Discover: Discovering important segments for classification of video
  events and recounting.
\newblock In {\em CVPR}, 2014.

\bibitem{tran2015learning}
D.~Tran, L.~Bourdev, R.~Fergus, L.~Torresani, and M.~Paluri.
\newblock Learning spatiotemporal features with {3D} convolutional networks.
\newblock In {\em ICCV}, 2015.

\bibitem{vahdat2013compositional}
A.~Vahdat, K.~Cannons, G.~Mori, S.~Oh, and I.~Kim.
\newblock Compositional models for video event detection: A multiple kernel
  learning latent variable approach.
\newblock In {\em ICCV}, 2013.

\bibitem{vedaldi2012efficient}
A.~Vedaldi and A.~Zisserman.
\newblock Efficient additive kernels via explicit feature maps.
\newblock {\em PAMI}, 34(3):480--492, 2012.

\bibitem{wang2013dense}
H.~Wang, A.~Kl{\"a}ser, C.~Schmid, and C.-L. Liu.
\newblock Dense trajectories and motion boundary descriptors for action
  recognition.
\newblock {\em IJCV}, 103(1):60--79, 2013.

\bibitem{wang2013action}
H.~Wang and C.~Schmid.
\newblock Action recognition with improved trajectories.
\newblock In {\em ICCV}, 2013.

\bibitem{dynamic_flow}
J.~Wang, A.~Cherian, and F.~Porikli.
\newblock Ordered pooling of optical flow sequences for action recognition.
\newblock {\em CoRR}, abs/1701.03246, 2017.

\bibitem{wang2015action}
L.~Wang, Y.~Qiao, and X.~Tang.
\newblock Action recognition with trajectory-pooled deep-convolutional
  descriptors.
\newblock In {\em CVPR}, 2015.

\bibitem{Wang2016}
L.~Wang, Y.~Xiong, Z.~Wang, Y.~Qiao, D.~Lin, X.~Tang, and L.~Van~Gool.
\newblock Temporal segment networks: Towards good practices for deep action
  recognition.
\newblock In {\em ECCV}, 2016.

\bibitem{wangtwo}
Y.~Wang, J.~Song, L.~Wang, L.~Van~Gool, and O.~Hilliges.
\newblock Two-stream sr-cnns for action recognition in videos.
\newblock In {\em BMVC}, 2016.

\bibitem{willems2009exemplar}
G.~Willems, J.~H. Becker, T.~Tuytelaars, and L.~J. Van~Gool.
\newblock Exemplar-based action recognition in video.
\newblock In {\em BMVC}, 2009.

\bibitem{wu2015deep}
J.~Wu, Y.~Yu, C.~Huang, and K.~Yu.
\newblock Deep multiple instance learning for image classification and
  auto-annotation.
\newblock In {\em CVPR}, 2015.

\bibitem{yi2016human}
Y.~Yi and M.~Lin.
\newblock Human action recognition with graph-based multiple-instance learning.
\newblock {\em Pattern Recognition}, 53:148--162, 2016.

\bibitem{yu2013propto}
F.~X. Yu, D.~Liu, S.~Kumar, T.~Jebara, and S.-F. Chang.
\newblock $propto $ svm for learning with label proportions.
\newblock {\em arXiv preprint arXiv:1306.0886}, 2013.

\bibitem{yue2015beyond}
J.~Yue-Hei~Ng, M.~Hausknecht, S.~Vijayanarasimhan, O.~Vinyals, R.~Monga, and
  G.~Toderici.
\newblock Beyond short snippets: Deep networks for video classification.
\newblock In {\em CVPR}, 2015.

\bibitem{zepeda2015exemplar}
J.~Zepeda and P.~Perez.
\newblock Exemplar svms as visual feature encoders.
\newblock In {\em CVPR}, 2015.

\bibitem{zhang2015self}
D.~Zhang, D.~Meng, C.~Li, L.~Jiang, Q.~Zhao, and J.~Han.
\newblock A self-paced multiple-instance learning framework for co-saliency
  detection.
\newblock In {\em ICCV}, 2015.

\end{thebibliography}
